\documentclass[conference]{IEEEtran}
\IEEEoverridecommandlockouts
\usepackage{cite}
\usepackage{booktabs}
\usepackage{amsmath,amssymb,amsfonts}
\usepackage{algorithmic}
\usepackage{graphicx}
\usepackage{textcomp}
\usepackage{hyperref}
\usepackage{fancyhdr}
\usepackage{enumitem}
\usepackage{tikz}
\usepackage{pgfplots}
\pgfplotsset{compat=1.18}
\usetikzlibrary{arrows.meta,positioning,shapes.geometric,calc}
\usetikzlibrary{decorations.pathreplacing}
\usepackage{xcolor}

\definecolor{atomhl}{HTML}{2980B9}
\definecolor{radius1}{HTML}{E74C3C}
\definecolor{bitone}{HTML}{2980B9}
\definecolor{bitzero}{HTML}{FFFFFF}
\definecolor{bondgray}{HTML}{2D3436}
\usepackage[ruled,vlined]{algorithm2e}
\SetAlgoCaptionSeparator{.}
\SetAlCapFnt{\small\bfseries}
\SetAlgoNlRelativeSize{-1}
\DontPrintSemicolon
\SetKwInOut{KwIn}{Input}
\SetKwInOut{KwOut}{Output}
\SetKwComment{Comment}{$\triangleright$ }{}
\usepackage[most]{tcolorbox}
\usepackage{caption}
\newtcolorbox{algobox}{
  enhanced, colback=gray!5, colframe=black!35,
  boxrule=0.6pt, arc=2mm, left=1mm,right=1mm,top=1mm,bottom=1mm
}
\definecolor{cB1}{HTML}{D9EAF7}
\definecolor{cB2}{HTML}{D5F5E3}
\definecolor{cB3}{HTML}{FDEBD0}
\definecolor{cB4}{HTML}{E8DAEF}
\definecolor{cB5}{HTML}{FADBD8}
\definecolor{cB6}{HTML}{EAECEE}
\definecolor{cSc}{HTML}{FFF2CC}

\usepackage{float}
\def\BibTeX{{\rm B\kern-.05em{\sc i\kern-.025em b}\kern-.08em
    T\kern-.1667em\lower.7ex\hbox{E}\kern-.125emX}}

\makeatletter
\newcommand{\linebreakand}{%
  \end{@IEEEauthorhalign}
  \hfill\mbox{}\par
  \mbox{}\hfill\begin{@IEEEauthorhalign}
}
\makeatother

\begin{document}

\title{Role-Aware Morgan Fingerprints for Reaction Yield Prediction}

\author{
\IEEEauthorblockN{Chinmay Mirji}
\IEEEauthorblockA{\textit{Aerospace Engineering}\\
\textit{Embry-Riddle Aeronautical University}\\
Daytona Beach, Florida\\
mirjic@my.erau.edu}
\thanks{The link to the repository can be accessed from here:
\url{https://github.com/chinmaymirji/morgan-fp-yield-prediction.git}}
\and
\IEEEauthorblockN{Prashant Shekhar}
\IEEEauthorblockA{\textit{Mathematics}\\
\textit{Embry-Riddle Aeronautical University}\\
Daytona Beach, Florida\\
shekharp@erau.edu}
\linebreakand
\IEEEauthorblockN{Foram Madiyar}
\IEEEauthorblockA{\textit{Chemistry}\\
\textit{Bethune Cookman University}\\
Daytona Beach, Florida\\
madiyarf@cookman.edu}
\and
\IEEEauthorblockN{Hao Peng}
\IEEEauthorblockA{\textit{Aerospace Engineering}\\
\textit{Embry-Riddle Aeronautical University}\\
Daytona Beach, Florida\\
pengh2@erau.edu}
}

\maketitle

\begin{abstract}
Predicting reaction yield from molecular structure and reaction
context can cut experimental trial-and-error and speed up
condition screening in synthetic chemistry. Recent methods for
this task use learned representations such as graph neural
networks or Transformer encoders over reaction SMILES (Simplified Molecular Input Line Entry System), but
these approaches carry heavy preprocessing overhead and can
break when input formatting is inconsistent. We propose MFP,
a reaction yield prediction method built on role-aware Morgan
fingerprints: count-based circular fingerprints are computed for
each reaction component, aggregated by chemical role (reactant,
reagent, product), and combined with transformation-sensitive
difference features into a fixed-length reaction descriptor fed
to a feed-forward neural regressor. We test MFP against
state of the art methods such as YieldBERT (with and without data augmentation) and GNAN (graph neural network) on
the Suzuki-Miyaura and Buchwald-Hartwig benchmarks using a
shared preprocessing and evaluation protocol. MFP reaches
$R^2 = 0.878$ on Suzuki-Miyaura and $R^2 = 0.969$ on
Buchwald-Hartwig while training an order of magnitude faster
than graph- or Transformer-based alternatives. A formal
complexity analysis confirms that MFP folds all representation
cost into a one-time preprocessing step, removing the per-epoch
message-passing overhead that graph methods carry. An ablation
over fingerprint radius and folded vector length shows that
radius-2 representations at \texttt{nBits}$\,=2048$ give the
best balance of accuracy, speed, and cross-split stability on
both datasets. These results establish MFP as an effective,
reproducible, and efficient baseline for reaction yield
prediction.
\end{abstract}


\section{Introduction}
Predicting how much product a chemical reaction will produce, given the participating components and experimental conditions, is a long-standing goal in computational chemistry. Reliable yield models can cut wet-lab workload by ranking catalyst, ligand, base, and solvent combinations and by flagging unpromising conditions before any experiment is run \cite{zuranski2021review}. Cross-coupling reactions, notably Suzuki-Miyaura \cite{perera2018suzuki_platform} and Buchwald-Hartwig \cite{ahneman2018buchwald_ml} couplings, have become standard benchmarks for this task because of their importance in pharmaceutical synthesis and the availability of curated high-throughput yield data.

How molecules are represented matters enormously. Extended-connectivity fingerprints (ECFPs), commonly called Morgan fingerprints, remain among the most popular molecular descriptors in cheminformatics because they efficiently encode local atomic neighborhoods and have proven useful across many property prediction tasks \cite{rogers2010ecfp}. Multiple large-scale studies have found that descriptor-based models can match or even outperform graph neural networks while requiring far less computation \cite{jiang2021gnn_vs_descriptors, yang2019chemprop}. A recent analysis of 25 pretrained molecular embedding models across 25 datasets reinforced this conclusion: nearly all learned representations performed at or below the level of baseline ECFP fingerprints \cite{praski2025benchmarking}.

At the same time, learned representations have gained traction. Message-passing networks over molecular graphs \cite{kwon2022gnan, han2024pretrained_gnn} and Transformer encoders applied to reaction SMILES (Simplified Molecular Input Line Entry System) \cite{schwaller2021yieldbert, schwaller2021rxnfp} can be effective, but their performance often hinges on preprocessing details: how separators are handled, whether whitespace is stripped, and whether component conventions are consistent. When the input encoding is fragile, failures can stem from formatting rather than chemistry.

Several recent analyses have also cautioned that good benchmark numbers do not guarantee real-world generalization. Saebi \textit{et al.} \cite{saebi2023realworld} showed that dataset bias, noise, and inconsistent measurement conditions can heavily distort reported metrics, and Voinarovska \textit{et al.} \cite{voinarovska2024challenges} catalogued persistent challenges that current methods leave unresolved. A deeper issue is that all existing approaches, whether descriptor-based, graph-based, or sequence-based, learn statistical correlations between molecular features and yield rather than the underlying causal mechanisms that govern reaction outcomes. No current model captures why a particular ligand-substrate pairing fails at oxidative addition or why a specific base promotes protodeboronation; they only learn that certain input patterns co-occur with low or high yields in the training data. This distinction matters because correlative models can break silently when deployed on reaction spaces whose structure-yield relationships differ from the training distribution. These warnings highlight the importance of transparent, reproducible benchmarking that weighs both accuracy and practical overhead.

This paper benchmarks learned and descriptor-based approaches under a single preprocessing and evaluation pipeline. We measure how close a lightweight baseline (Morgan fingerprints + ANN) can get to a strong GNN baseline (GNAN) \cite{kwon2022gnan} while using far less preprocessing and compute. Beyond this cross-method comparison, we run a systematic ablation over Morgan fingerprint radius and folded vector length to find the configuration that best balances accuracy, speed, and stability across splits.

We make four contributions:
\begin{enumerate}[label=\roman*.]
\item We introduce MFP, a role-aware Morgan fingerprint
reaction descriptor that splits each reaction into six
fingerprint blocks capturing component identity, structural
transformation, and transformation magnitude, appended with
scalar features to form a fixed-length input for a lightweight
MLP regressor.
\item We test MFP against two families of learned
representations (reaction Transformers and graph neural
networks) under a shared pipeline and show that it approaches
GNAN accuracy at a fraction of the training cost. A formal
complexity analysis pins down where the cost difference
originates.
\item We ablate fingerprint radius and folded vector length
across both datasets, finding that radius-2 at
\texttt{nBits}$\,=2048$ is the strongest default. We ground
this finding in cross-coupling chemistry: radius-2
neighborhoods capture the ortho substitution and ligand steric
environments that most influence yield.
\item We release all code, preprocessing scripts, and trained
models to support reproducible comparisons.
\end{enumerate}

\section{Related Work}

\paragraph{Descriptor-based approaches.}
Extended-connectivity fingerprints (ECFPs), also called Morgan fingerprints, remain among the most widely used molecular representations. They encode local atomic environments efficiently and retain strong predictive utility across cheminformatics tasks \cite{rogers2010ecfp}. Their speed and reproducibility make them a natural starting point for yield modeling. Head-to-head comparisons between descriptors and graph models have repeatedly shown that fingerprint-based methods match or outperform learned representations on diverse molecular property endpoints \cite{jiang2021gnn_vs_descriptors, yang2019chemprop, praski2025benchmarking}. Other lightweight descriptors exist as well: the differential reaction fingerprint (DRFP) \cite{probst2022drfp} captures the transformation between reactants and products in a compact form and has shown competitive performance for both classification and yield prediction.

\paragraph{Early ML-driven yield prediction.}
The value of supervised learning for yield prediction was first demonstrated by Ahneman \textit{et al.} \cite{ahneman2018buchwald_ml} on Buchwald-Hartwig C-N cross-coupling, using experimentally curated high-throughput data. Around the same time, Perera \textit{et al.} \cite{perera2018suzuki_platform} developed an automated nanomole-scale screening platform that produced the Suzuki-Miyaura datasets now commonly used as benchmarks. Both datasets have since become standard testbeds.

\paragraph{Learned reaction representations.}
Transformer-based methods treat reaction SMILES as sequences and fine-tune pretrained language models for regression. YieldBERT \cite{schwaller2021yieldbert} showed that yield can be predicted effectively from reaction strings, especially when the encoder is pretrained on reaction data. The randomized SMILES enumeration technique introduced by Bjerrum \cite{bjerrum2017smiles} has become a standard augmentation strategy for such models. The RXNFP architecture \cite{schwaller2021rxnfp} further demonstrated that Transformer encoders can learn chemically meaningful embeddings directly from reaction text. More recently, Shi \textit{et al.} \cite{shi2024reamvp} proposed ReaMVP, which incorporates multi-view pretraining with 3D geometry to push state-of-the-art performance, particularly on out-of-sample data. Wang \textit{et al.} \cite{yin2024enhancing} applied contrastive learning to yield prediction, using reaction-condition-based pretraining to improve generalization across reaction categories.

\paragraph{Graph-based approaches.}
Graph neural networks model molecular topology explicitly. Kwon \textit{et al.} \cite{kwon2022gnan} proposed GNAN, a framework with integrated uncertainty quantification that achieved strong results on multiple benchmarks. Srinivas \textit{et al.} \cite{singh2024deep} combined neural feature extraction with Gaussian processes in a deep kernel learning framework, obtaining comparable accuracy with calibrated uncertainty estimates suited for Bayesian optimization. Han \textit{et al.} \cite{han2024pretrained_gnn} later showed that pre-trained graph representations can improve both accuracy and cross-domain transferability. These graph-based models capture rich structural interactions but typically need more involved preprocessing and graph construction than descriptor-based alternatives.

\paragraph{Challenges and critiques.}
\.{Z}ura\'{n}ski \textit{et al.} \cite{zuranski2021review} reviewed supervised learning for reaction yields and stressed the importance of dataset quality, representation choice, and experimental design. Saebi \textit{et al.} \cite{saebi2023realworld} showed that dataset bias, measurement noise, and inconsistent conditions can heavily influence reported performance. Voinarovska \textit{et al.} \cite{voinarovska2024challenges} argued that favorable benchmark numbers do not necessarily translate to robust real-world deployment. Ma \textit{et al.} \cite{ma2024we} reframed yield prediction as an imbalanced regression problem, pointing out that models often fit well in the data-rich low-yield region while underperforming on the high-yield reactions that matter most to chemists.

Building on these findings, we compare Transformer-, graph-, and fingerprint-based model families under a unified protocol, assessing not only accuracy but also the tradeoff between complexity, cost, and reproducibility.

\section{Datasets and Preprocessing}

\subsection{Data and Code Availability}
The datasets, pretrained model checkpoints, and all source code
used in this study are publicly available. For reaction SMILES
and yield labels, we use the resources released alongside
YieldBERT by Schwaller \textit{et al.} \cite{schwaller2021yieldbert}
at \url{https://github.com/rxn4chemistry/rxn_yields/tree/master}
and the Buchwald-Hartwig dataset from the Doyle Lab
\cite{ahneman2018buchwald_ml} at
\url{https://github.com/doylelab/rxnpredict/tree/master}.

\subsection{Suzuki-Miyaura Reaction Yield Dataset}
The Suzuki-Miyaura dataset comes from a high-throughput nanomole-scale cross-coupling screening platform \cite{perera2018suzuki_platform}. Each split file contains $N=5760$ reactions with two fields: \texttt{rxn}, a reaction SMILES-like string encoding the participating components, and \texttt{y}, the corresponding yield on a normalized $[0,1]$ scale. We use a fixed $70\%/30\%$ train-test partition for each split, applied consistently across all random split permutations.

\begin{figure*}[!t]
    \centering
    \includegraphics[width=1\linewidth]{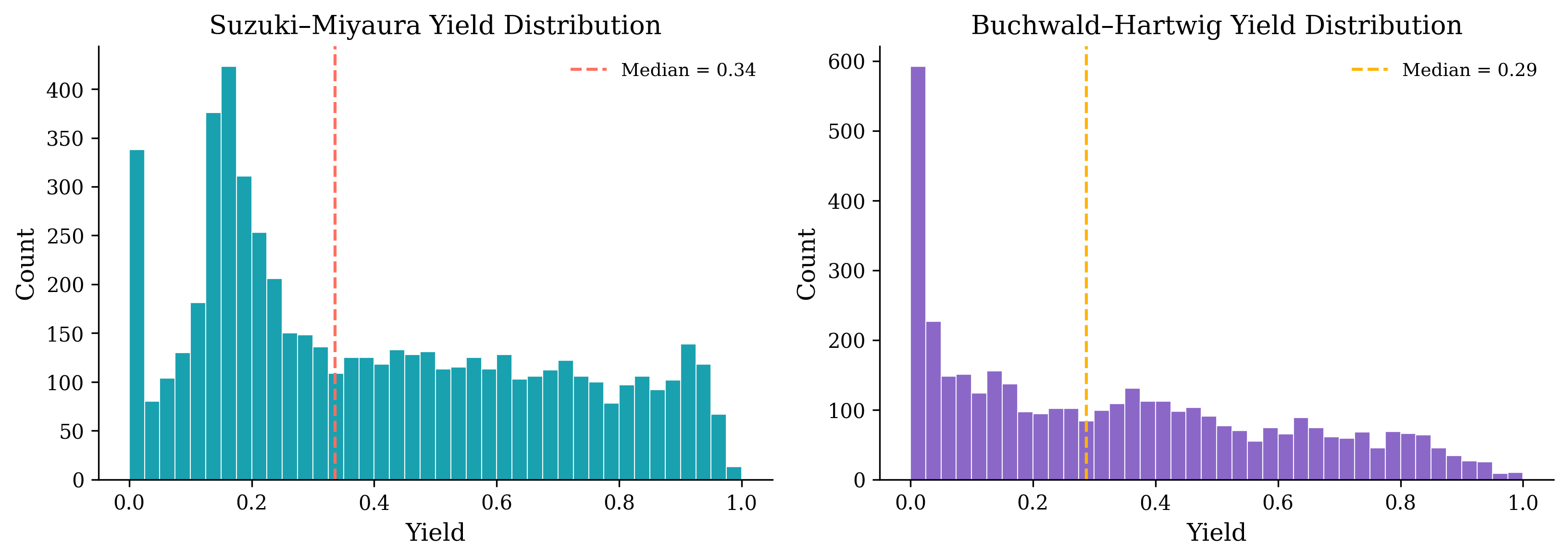}
    \caption{Yield Histograms}
    \label{fig:yield_hist}
\end{figure*}

\begin{figure*}[!t]
    \centering
    \includegraphics[width=1\linewidth]{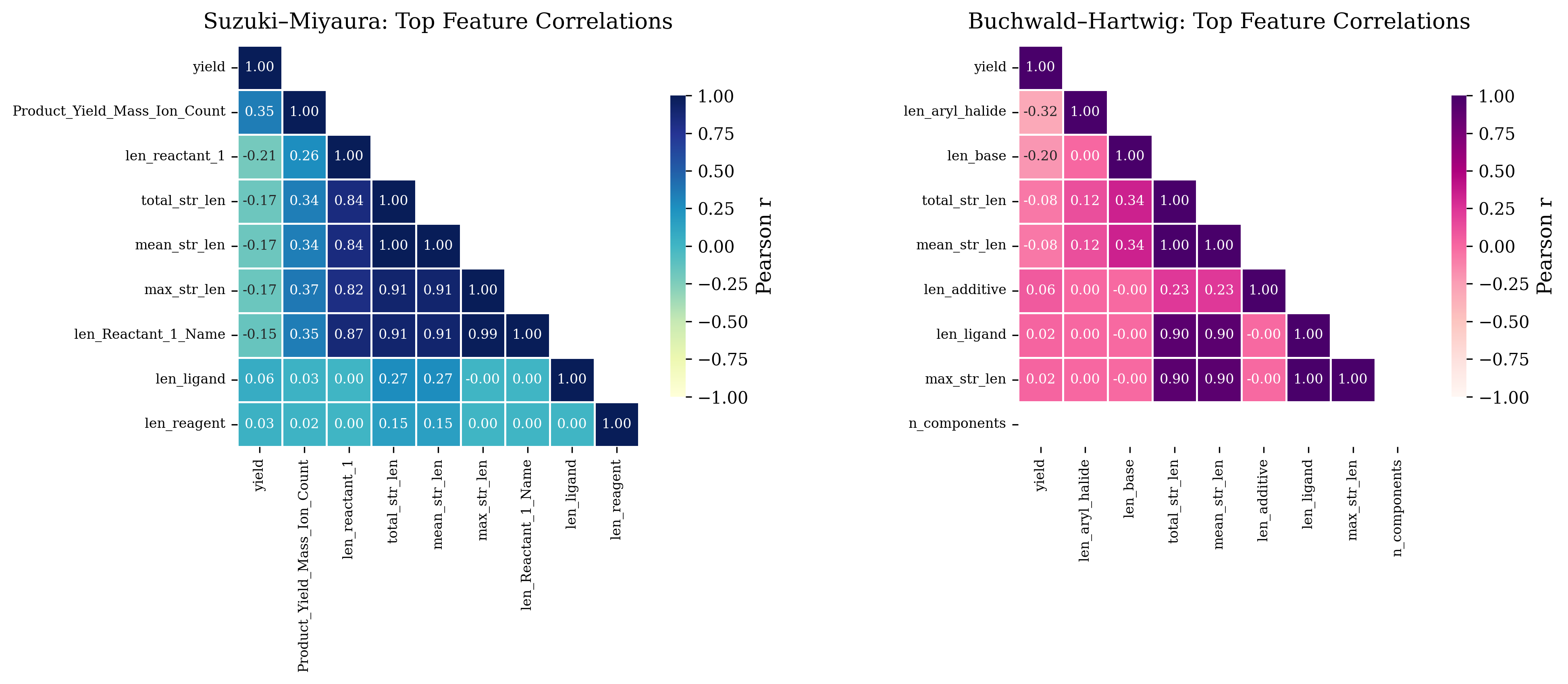}
    \caption{Correlation Heatmaps}
    \label{fig:corr}
\end{figure*}

\subsection{Buchwald-Hartwig Reaction Yield Dataset}
We also evaluate on a Buchwald-Hartwig C-N cross-coupling dataset collected via high-throughput experimentation \cite{ahneman2018buchwald_ml}. Each entry corresponds to a reaction performed in a controlled combinatorial condition space and labeled with the observed product yield. Reactions are encoded as SMILES strings of the form $\textit{reactants}>\textit{reagents}>\textit{products}$, with yields scaled to $[0,1]$. This dataset has become a widely used benchmark for yield prediction \cite{zuranski2021review, saebi2023realworld}.

\subsection{Exploratory Data Analysis}
Figure~\ref{fig:yield_hist} shows the yield distributions for both datasets. The Suzuki-Miyaura distribution is broadly spread across $[0,1]$ with a median of $0.34$, while the Buchwald-Hartwig distribution is heavily concentrated in the low-yield region (median $0.29$) with a sharp spike near zero. Both datasets are dominated by low-yield reactions, a pattern that creates challenges for regression models \cite{ma2024we}.

To check whether yield can be trivially predicted from string-level features, we compute simple statistics (component string lengths, total character counts, component counts) and examine their correlations with yield. Figure~\ref{fig:corr} shows these correlations for both datasets. Even the strongest associations are weak ($|r| < 0.35$ for Suzuki-Miyaura; $|r| < 0.25$ for Buchwald-Hartwig), confirming that string-level properties carry little predictive information \cite{saebi2023realworld}. Predictive power must therefore come from chemical identity and reaction context, not superficial input features.

The zero-yield spike in both datasets is not a measurement artifact. It reflects genuine reaction failure modes documented in high-throughput cross-coupling studies. In the Suzuki-Miyaura system, near-zero yields commonly result from catalyst poisoning, protodeboronation of the boronic acid partner (cleavage of the C-B bond by solvent or base before transmetalation), or homocoupling byproducts that consume the aryl halide \cite{perera2018suzuki_platform}. In the Buchwald-Hartwig system, failures are most often traced to unproductive ligand-palladium combinations that stall at oxidative addition, competitive reduction of the aryl halide to the parent arene, or base-mediated decomposition of the amine nucleophile \cite{ahneman2018buchwald_ml}. Because combinatorial screening tests every catalyst-ligand-base combination regardless of chemical compatibility, these failure modes are overrepresented. The resulting imbalance means that models can achieve strong aggregate metrics simply by fitting the abundant low-yield region well, while performing poorly on the sparser high-yield reactions that are more practically relevant \cite{ma2024we}.

\subsection{Reaction String Normalization}
The raw reaction strings sometimes contain non-standard separators that disrupt tokenization, which is especially problematic for Transformer-based models \cite{schwaller2021yieldbert}. We therefore apply a deterministic normalization to every reaction string before training and before any augmentation:
\begin{enumerate}[label=\roman*.]
  \item Remove whitespace and pipe characters (\texttt{|}).
  \item Replace \texttt{\string~} with the fragment separator \texttt{.}.
  \item Collapse repeated separators (e.g., \texttt{..} $\rightarrow$ \texttt{.}).
\end{enumerate}
This normalization stabilizes SMILES-based models and ensures consistent formatting across all methods.

\section{Methodology}

\subsection{Dataset, Split Protocol, and Targets}
We evaluate all models using ten random splits, following standard practice \cite{schwaller2021yieldbert, kwon2022gnan}. Each split file contains reaction strings \texttt{rxn} and yields \texttt{y} in $[0,1]$. For Suzuki-Miyaura, each split uses $N_{\mathrm{train}}=4032$ and $N_{\mathrm{test}}=1728$. For Buchwald-Hartwig, $N_{\mathrm{train}}=2767$ and $N_{\mathrm{test}}=830$. We report per-split metrics and mean $\pm$ standard deviation across splits.

\subsection{Per-Split Label Standardization and Inverse Transform}
All methods are trained as regressors. For split $i$, we compute the training mean $\mu_i$ and standard deviation $\sigma_i$ from training yields only, then standardize:
\[
y_{\mathrm{scaled}} = \frac{y - \mu_i}{\sigma_i}.
\]
Predictions $\hat{y}_{\mathrm{scaled}}$ are mapped back for evaluation:
\[
\hat{y} = \hat{y}_{\mathrm{scaled}}\sigma_i + \mu_i.
\]
All reported MAE, RMSE, and $R^2$ values use the original yield scale.

\subsection{Model Families}

\subsubsection{YieldBERT (Reaction Transformer Regression)}
YieldBERT \cite{schwaller2021yieldbert} treats the reaction string as a sequence, fine-tuning a pretrained BERT-style reaction encoder with a regression head. Each normalized reaction string is tokenized and passed through the Transformer; a pooled representation is mapped to a scalar $\hat{y}_{\mathrm{scaled}}$. Training minimizes MSE on standardized labels. The encoder leverages reaction-aware pretraining \cite{schwaller2021rxnfp}, which produces chemically meaningful latent representations.

\subsubsection{YieldBERT-DA (Augmentation + Optional Test-Time Averaging)}
YieldBERT-DA adds string-level data augmentation to enforce invariance to equivalent SMILES forms and component orderings, building on the randomized SMILES technique of Bjerrum \cite{bjerrum2017smiles}:
\begin{enumerate}[label=\roman*.]
  \item Randomized SMILES: RDKit SMILES randomization ({$doRandom=True$}) generates equivalent non-canonical representations.
  \item Component shuffling: the order of \texttt{.}-separated components within each reaction side is randomly permuted.
\end{enumerate}
Importantly, neither operation alters the underlying chemistry. Randomized SMILES produces a different character string for the same molecule: the atom ordering and branch notation change, but the molecular graph, bond connectivity, and stereochemistry remain identical. Component shuffling reorders entries within a reaction side without adding, removing, or modifying any species. The augmentation therefore teaches the model that these surface-level variations are uninformative, without introducing any synthetic reactions that lack experimental grounding.
At training time, each reaction is expanded by a factor $K$. At test time, we optionally apply test-time augmentation (TTA): generate $N_{\mathrm{TTA}}$ variants per test reaction, predict each, and average.

Let $r$ denote the normalized reaction string and $\tau(r)=(w_1,\dots,w_T)$ its token sequence. The Transformer encoder produces:
\[
H = \mathrm{BERT}_{\theta}(\tau(r)) \in \mathbb{R}^{T\times d_h}.
\]
With pooled representation $h = \mathrm{pool}(H)$, the regression head outputs:
\[
\hat{y}_s = g_{\theta}(h), \qquad \hat{y} = \hat{y}_s \sigma + \mu,
\]
trained with MSE:
\[
\mathcal{L}(\theta) = \frac{1}{N}\sum_{i=1}^{N}\left(\hat{y}_{s,i}-y_{s,i}\right)^2.
\]

Let $\mathcal{A}(r)$ be an augmentation operator producing an equivalent reaction string. During training, each sample expands to $\{\tilde{r}_{i,j}\}_{j=1}^{K}$ sharing label $y_{s,i}$. At test time, TTA averages over $N_{\mathrm{TTA}}$ variants:
\[
\hat{y}_s(r) = \frac{1}{N_{\mathrm{TTA}}}\sum_{j=1}^{N_{\mathrm{TTA}}} g_{\theta}\!\left(\mathrm{pool}\left(\mathrm{BERT}_{\theta}(\tau(\tilde{r}_j))\right)\right).
\]

\subsubsection{GNAN (Graph Neural Network with Uncertainty Awareness)}
GNAN \cite{kwon2022gnan} represents each reaction as a set of molecular graphs and uses message-passing neural networks to encode atom- and bond-level structure. It incorporates heteroscedastic regression and category-wise regularization to produce calibrated uncertainty estimates alongside yield predictions. GNAN has shown strong performance on multiple yield benchmarks and represents a high-accuracy reference for graph-based methods. Related work by Srinivas \textit{et al.} \cite{singh2024deep} paired neural feature learning with Gaussian processes, achieving comparable accuracy with calibrated uncertainty suited for Bayesian optimization. Han \textit{et al.} \cite{han2024pretrained_gnn} later improved accuracy and transferability by incorporating pre-trained graph representations.

\begin{figure}[!ht]
    \centering
    \includegraphics[width=0.8\linewidth]{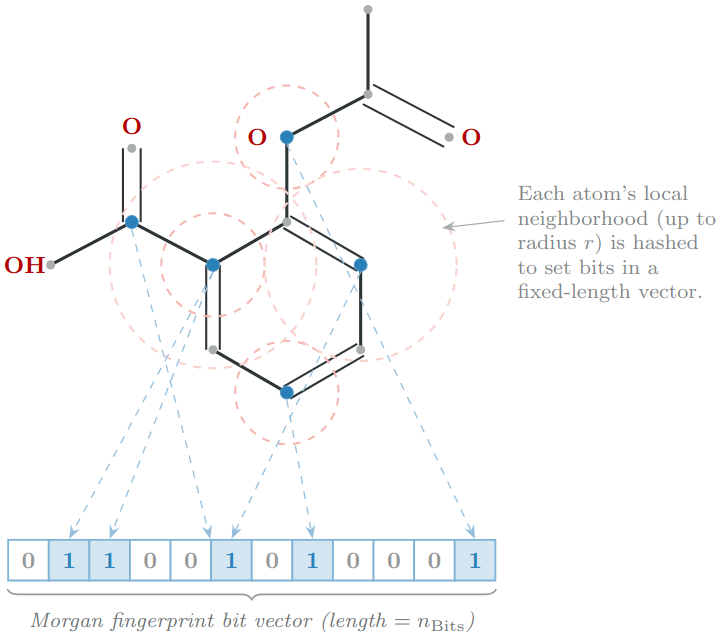}
    \caption{Schematic illustration of Morgan Fingerprints generation}
    \label{fig:fp_rep}
\end{figure}

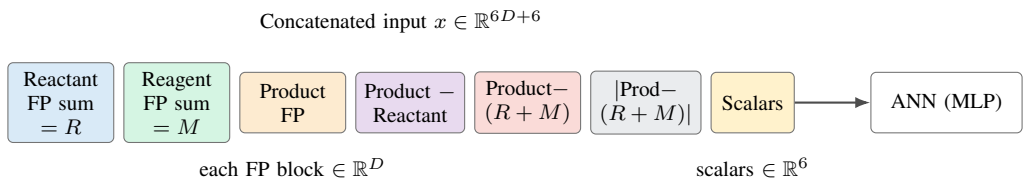
\begin{figure*}[!h]
\centering
\begin{tikzpicture}[
  font=\footnotesize,
  seg/.style={draw=black!55, rounded corners=2pt, minimum height=8mm, align=center},
  arrow/.style={-{Latex[length=2mm]}, draw=black!70, line width=0.8pt}
]
\node[seg, fill=cB1, minimum width=14mm] (b1) {Reactant\\FP sum\\$= R$};
\node[seg, fill=cB2, minimum width=14mm, right=1.2mm of b1] (b2) {Reagent\\FP sum \\$=M$};
\node[seg, fill=cB3, minimum width=14mm, right=1.2mm of b2] (b3) {Product\\FP};
\node[seg, fill=cB4, minimum width=14mm, right=1.2mm of b3] (b4) {Product $-$\\Reactant};
\node[seg, fill=cB5, minimum width=14mm, right=1.2mm of b4] (b5) {Product$-$\\$(R+M)$};
\node[seg, fill=cB6, minimum width=14mm, right=1.2mm of b5] (b6) {$|$Prod$-$\\$(R+M)|$};
\node[seg, fill=cSc, minimum width=11mm, right=1.2mm of b6] (sc) {Scalars};

\coordinate (mid) at ($(b1.west)!0.5!(sc.east)$);
\node[above=8mm of mid] {Concatenated input $x \in \mathbb{R}^{6D+6}$};

\node[below=2mm of b3, align=center] {each FP block $\in \mathbb{R}^{D}$};
\node[below=2mm of sc, align=center] {scalars $\in \mathbb{R}^{6}$};

\node[seg, minimum width=20mm, fill=white, right=10mm of sc] (mlp) {ANN (MLP)};
\draw[arrow] (sc.east) -- (mlp.west);

\end{tikzpicture}
\caption{Reaction feature vector for the Morgan fingerprint + ANN model. Six fingerprint blocks (each length $D$) are concatenated with a small set of scalar descriptors and passed to an MLP regressor.}
\label{fig:morgan_blocks}
\end{figure*}

\subsubsection{Morgan Fingerprints + ANN (Descriptor Baseline)}
Morgan (circular) fingerprints \cite{rogers2010ecfp} represent molecular structure at fixed length by hashing local atom-centered neighborhoods up to a chosen radius. Figure~\ref{fig:fp_rep} illustrates the embedding. We use \emph{count}-based Morgan fingerprints from RDKit, apply light compression (count clipping followed by $\log(1+x)$), and build a reaction descriptor through \emph{role-wise aggregation}: reactants (left side) are summed separately from reagents (middle side). We then concatenate these role-wise aggregates with product-centered difference features and a small set of scalar descriptors, and train a lightweight MLP regressor on standardized yields. The design draws on the differential representation idea behind DRFP \cite{probst2022drfp}, adapted here for count-based fingerprints with explicit role separation.

Let $\phi(\cdot)\in\mathbb{R}^{D}$ denote the compressed Morgan \emph{count} fingerprint for a molecule, formed by concatenating fingerprints across multiple radii and bit-sizes into a single length-$D$ vector. For a reaction with reactant set $L$, reagent set $M$, and product $p$, we compute role-wise sums $R=\sum_{s\in L}\phi(s)$ and $M=\sum_{s\in M}\phi(s)$, along with the product fingerprint $P=\phi(p)$. The full reaction descriptor (Figure~\ref{fig:morgan_blocks}) concatenates six fingerprint blocks: the reactant sum $R$, the reagent sum $M$, the product fingerprint $P$, a reactant-to-product difference $(P-R)$, a full-reaction difference $P-(R+M)$, and its element-wise absolute value $|P-(R+M)|$. The first three blocks encode what is present in each role. The difference $P - R$ captures the net structural change from the coupling step, while $P - (R+M)$ measures what changes after accounting for both reactants and reagents. The absolute value $|P-(R+M)|$ retains the magnitude of change regardless of sign, which helps when both gains and losses of a substructure carry predictive weight. Together, these six blocks give the model complementary views of the reaction: identity, transformation, and transformation magnitude. A small set of scalar features $s\in\mathbb{R}^{6}$ (component counts and presence flags) is appended, yielding a final input $x\in\mathbb{R}^{6D+6}$ fed to an MLP that predicts standardized yield $\hat{y}_s=f_\theta(x)$, evaluated after inverse transform $\hat{y}=\hat{y}_s\sigma+\mu$.

\subsection{Training, Metrics, and Diagnostic Plots}
All models are trained on each split independently using standardized labels and MSE loss. We evaluate using
\[
\mathrm{MAE},\ \mathrm{RMSE},\ R^2
\]
computed on unscaled yields in $[0,1]$. We generate measured-vs.-predicted scatter plots for each split and an aggregate plot over all test predictions. As a sanity check, we also compute a constant baseline that predicts the training mean $\mu_i$ for every test sample in split $i$.

\section{Experiments}

\subsection{Unified Evaluation Protocol}
All methods share the same reaction-string normalization, per-split label standardization, and train/test partitioning. For augmented Transformer runs, augmentation applies only to training data (and optionally to test data for TTA). For fingerprint and graph models, reaction roles stay fixed and no operation leaks information across partitions. This shared protocol ensures that performance differences reflect genuine representational and architectural distinctions, not preprocessing inconsistencies \cite{saebi2023realworld}.

\subsection{Computational Efficiency and Preprocessing Cost}
We aim to quantify the accuracy-efficiency tradeoff across model families. We do not claim that Morgan fingerprints beat GNAN in accuracy; instead, we measure how close a lightweight baseline can get while using far less overhead. We compare (i) preprocessing burden, (ii) feature construction time per reaction, and (iii) training time per split on the same hardware. Morgan fingerprints need only SMILES validation and fingerprint computation \cite{rogers2010ecfp}. Graph pipelines additionally require reaction-graph construction \cite{kwon2022gnan}, and Transformer pipelines require tokenization-sensitive string processing \cite{schwaller2021yieldbert}.

\subsection{Morgan Fingerprint Ablation Design}
\label{sec:ablation_design}
To characterize how sensitive the Morgan + ANN baseline is to its key hyperparameters, we ablate over fingerprint radius and folded vector length (\texttt{nBits}). We test eight configurations spanning three radius settings: radius~2 only, radius~3 only, and the concatenation of radii~2 and~3. For each, we vary \texttt{nBits} across $\{1024, 2048, 4096\}$ (single-radius) or $\{(2048,2048), (4096,4096)\}$ (concatenated). This grid separates the effect of neighborhood depth (radius) from hash collision frequency (\texttt{nBits}).

We first screen all configurations on split~0 of each dataset to identify promising settings, then run the best-performing radius across all ten splits to assess stability. All other hyperparameters are held fixed so that differences are attributable to the fingerprint representation alone.

\section{Results and Discussion}

\subsection{Cross-Method Predictive Performance}
Table~\ref{tab:results} summarizes performance on both benchmarks. The comparison is not meant to show that Morgan fingerprints beat GNAN. Rather, it shows that a lightweight descriptor baseline can reach comparable performance with much less preprocessing and compute. Morgan fingerprints \cite{rogers2010ecfp} sidestep the tokenization issues that affect Transformer models \cite{schwaller2021yieldbert} and skip the graph construction needed by GNNs \cite{kwon2022gnan}, while still capturing informative local chemical environments. For reference, recent state-of-the-art methods like ReaMVP \cite{shi2024reamvp} have pushed further by incorporating 3D geometry, but at the cost of much greater architectural complexity.

\begin{table*}[!t]
    \centering
    \begin{tabular}{lcccccc}\toprule
         & \multicolumn{3}{c}{Buchwald-Hartwig} & \multicolumn{3}{c}{Suzuki-Miyaura} \\\midrule
       \textbf{Method} & \textbf{MAE (\%p)} & \textbf{RMSE (\%p)} & \textbf{$R^{2}$} & \textbf{MAE (\%p)} & \textbf{RMSE (\%p)} & \textbf{$R^{2}$} \\\midrule
 YieldBERT \cite{schwaller2021yieldbert} & $3.990 \pm 0.153$ & $6.014 \pm 0.272$ & $0.951 \pm 0.005$ & $8.128 \pm 0.344$ & $12.073 \pm 0.463$ & $0.815 \pm 0.013$ \\
 YieldBERT-DA \cite{schwaller2021yieldbert} & $3.090 \pm 0.118$ & $4.799 \pm 0.261$ & $0.969 \pm 0.004$ & $6.598 \pm 0.270$ & $10.524 \pm 0.482$ & $0.859 \pm 0.012$ \\
 GNAN \cite{kwon2022gnan} & $2.884 \pm 0.058$ & $4.413 \pm 0.108$ & $0.974 \pm 0.001$ & $5.962 \pm 0.230$ & $9.373 \pm 0.484$ & $0.888 \pm 0.010$ \\
 \textbf{ANN + Morgan FP} \cite{rogers2010ecfp} & $\mathbf{3.239 \pm 0.100}$ & $\mathbf{4.795 \pm 0.175}$ & $\mathbf{0.969 \pm 0.003}$ & $\mathbf{6.452 \pm 0.202}$ & $\mathbf{9.790 \pm 0.411}$ & $\mathbf{0.878 \pm 0.009}$ \\
 \bottomrule
    \end{tabular}
    \caption{Comparison of yield prediction performance. All metrics are reported as mean $\pm$ standard deviation across ten random splits. MAE and RMSE are in percentage points (\%p). Best lightweight baseline results are in \textbf{bold}.}
    \label{tab:results}
\end{table*}

\subsection{Accuracy-Efficiency Tradeoff}

GNAN achieves the highest accuracy on both datasets, as Kwon \textit{et al.} \cite{kwon2022gnan} and Han \textit{et al.} \cite{han2024pretrained_gnn} also reported. The Morgan fingerprint + ANN baseline, however, reaches a competitive range (especially on Suzuki-Miyaura) at a fraction of the computational cost. A brief complexity analysis shows why.

\subsubsection{Inference Cost}
Consider a reaction involving $K$ molecules with mean atom count $\bar{n}$ and average bond degree $d$. For the Morgan model with radius $r$, folded vector length $b$, and MLP hidden width $h$, the per-reaction cost is:
\begin{equation}
C_{\mathrm{Morgan}} = O\!\left(K\bar{n}\, d^{\,r}\right) + O(b \cdot h),
\label{eq:cost_morgan}
\end{equation}
where the first term is a \emph{fixed hash} with no learnable parameters and the second is the MLP forward pass. For GNAN with $L$ message-passing layers and hidden dimension $h_g$:
\begin{equation}
C_{\mathrm{GNAN}} = O\!\left(L \cdot K\bar{n}\, d \cdot h_g^{\,2}\right).
\label{eq:cost_gnan}
\end{equation}
The $L \cdot h_g^{\,2}$ factor reflects the learned weight matrices applied at every atom in every layer. The fingerprint model avoids this entirely by replacing learned message passing with a deterministic hash.

\subsubsection{Training Cost}
The gap grows during training. Morgan fingerprints are computed once and cached, so the optimizer only updates the small MLP on the precomputed feature matrix. Over $E$ epochs and $N$ reactions:
\begin{equation}
T_{\mathrm{Morgan}}^{\,\mathrm{train}} = O\!\left(N \cdot \bar{n}\, d^{\,r}\right) + E \cdot O(N \cdot b \cdot h).
\label{eq:train_morgan}
\end{equation}
GNAN runs the full message-passing forward pass and backpropagates through all $L$ layers at every step:
\begin{equation}
T_{\mathrm{GNAN}}^{\,\mathrm{train}} = E \cdot O\!\left(N \cdot L \cdot \bar{n}\, d \cdot h_g^{\,2}\right).
\label{eq:train_gnan}
\end{equation}
Amortizing representation cost into a one-time preprocessing step is a structural advantage that graph-based architectures cannot replicate without sacrificing end-to-end differentiability.

With the benchmark settings ($r\!=\!2$, $b\!=\!2048$, $h\!=\!256$ for Morgan; $L\!=\!3$ message-passing steps with a 512-dimensional prediction head for GNAN \cite{kwon2022gnan}), the difference shows up clearly in the training times reported in Tables~\ref{tab:sm_fp_ablation_split0} and~\ref{tab:bh_fp_ablation_split0}. In practice, the fingerprint baseline is a good fit for rapid experimentation: running hyperparameter sweeps, ablation studies, or establishing a reference score before investing in heavier architectures \cite{zuranski2021review, probst2022drfp, jiang2021gnn_vs_descriptors, praski2025benchmarking}.

\subsection{Morgan Fingerprint Ablation}
\label{sec:ablation_results}

Having established that the Morgan + ANN baseline is competitive, we now examine how radius and \texttt{nBits} affect performance, training cost, and cross-split stability.

\subsubsection{Single-Split Screening (Split 0)}
Tables~\ref{tab:sm_fp_ablation_split0} and~\ref{tab:bh_fp_ablation_split0} report ablation results on split~0. Figures~\ref{fig:sm_r2_vs_nbits_concat} and~\ref{fig:bh_r2_vs_nbits_concat} plot $R^2$ against \texttt{nBits} for the three radius settings.

On Suzuki-Miyaura, radius-2 fingerprints consistently outperform radius-3 across all \texttt{nBits} values, with the best single-split result ($R^2 = 0.845$, RMSE $= 0.110$) at radius~2 with \texttt{nBits}$\,=4096$. Going from radius~2 to 3 degrades accuracy modestly, suggesting that the broader neighborhoods introduce hash collisions or noise that offset any additional structural information. Concatenating radii~2+3 does not improve over radius~2 alone and costs more training time, indicating that naively combining radii adds no complementary signal for this dataset.

On Buchwald-Hartwig the picture is similar: radius-2 achieves the highest $R^2$, with the best result ($R^2 = 0.948$) at \texttt{nBits}$\,=1024$. The Buchwald-Hartwig dataset is less sensitive to \texttt{nBits} within the radius-2 family (all three settings reach $R^2 \geq 0.946$), likely because its combinatorial reaction space is more tightly structured \cite{ahneman2018buchwald_ml}. As with Suzuki-Miyaura, concatenated configurations underperform single-radius ones while taking longer to train.

\begin{table*}
\centering
\begin{minipage}[t]{0.48\textwidth}
\centering
\small
\begin{tabular}{lccccc}
\toprule
\textbf{Config} & \textbf{Radii} & \textbf{nBits} & \textbf{$R^2$} & \textbf{RMSE} & \textbf{Train (s)} \\
\midrule
r2\_nb1024   & 2     & 1024        & 0.830 & 0.115 & 55.2 \\
r2\_nb2048   & 2     & 2048        & 0.836 & 0.113 & 201.4 \\
r2\_nb4096   & 2     & 4096        & \textbf{0.845} & \textbf{0.110} & 522.1 \\
r3\_nb1024   & 3     & 1024        & 0.817 & 0.119 & 62.7 \\
r3\_nb2048   & 3     & 2048        & 0.829 & 0.115 & 190.7 \\
r3\_nb4096   & 3     & 4096        & 0.811 & 0.121 & 193.8 \\
r23\_nb2048  & 2,3   & 2048, 2048  & 0.821 & 0.118 & 409.1 \\
r23\_nb4096  & 2,3   & 4096, 4096  & 0.811 & 0.121 & 594.4 \\
\bottomrule
\end{tabular}
\captionof{table}{Morgan FP ablation on Suzuki-Miyaura split~0.}
\label{tab:sm_fp_ablation_split0}
\end{minipage}\hfill
\begin{minipage}[t]{0.48\textwidth}
\centering
\small
\begin{tabular}{lccccc}
\toprule
\textbf{Config} & \textbf{Radii} & \textbf{nBits} & \textbf{$R^2$} & \textbf{RMSE} & \textbf{Train (s)} \\
\midrule
r2\_nb1024   & 2     & 1024        & \textbf{0.948} & \textbf{6.260} & 41.7 \\
r2\_nb2048   & 2     & 2048        & 0.946 & 6.391 & 89.6 \\
r2\_nb4096   & 2     & 4096        & 0.946 & 6.387 & 118.5 \\
r3\_nb1024   & 3     & 1024        & 0.942 & 6.658 & 33.8 \\
r3\_nb2048   & 3     & 2048        & 0.948 & 6.271 & 201.3 \\
r3\_nb4096   & 3     & 4096        & 0.938 & 6.866 & 153.7 \\
r23\_nb2048  & 2,3   & 2048, 2048  & 0.933 & 7.122 & 153.5 \\
r23\_nb4096  & 2,3   & 4096, 4096  & 0.930 & 7.275 & 801.8 \\
\bottomrule
\end{tabular}
\captionof{table}{Morgan FP ablation on Buchwald-Hartwig split~0.}
\label{tab:bh_fp_ablation_split0}
\end{minipage}
\end{table*}

\begin{figure*}[!t]
\centering
\begin{minipage}[t]{0.48\textwidth}
\centering
\begin{tikzpicture}
\begin{axis}[
    width=\linewidth, height=5.5cm,
    xlabel={\texttt{nBits}}, ylabel={$R^2$ (split 0)},
    xmode=log, log basis x=2,
    xtick={1024,2048,4096,8192}, xticklabels={1024,2048,4096,8192},
    ymin=0.80, ymax=0.85, grid=both,
    legend style={at={(0.02,0.02)},anchor=south west,font=\scriptsize},
    tick label style={font=\small}, label style={font=\small},
]
\addplot+[mark=*, thick] coordinates {(1024,0.829923) (2048,0.836096) (4096,0.845056)};
\addlegendentry{Radius 2}
\addplot+[mark=square*, thick] coordinates {(1024,0.816648) (2048,0.828581) (4096,0.811263)};
\addlegendentry{Radius 3}
\addplot+[mark=triangle*, thick] coordinates {(4096,0.821443) (8192,0.810805)};
\addlegendentry{Radii 2+3}
\end{axis}
\end{tikzpicture}
\captionof{figure}{Suzuki-Miyaura split~0: $R^2$ vs.\ \texttt{nBits}.}
\label{fig:sm_r2_vs_nbits_concat}
\end{minipage}\hfill
\begin{minipage}[t]{0.48\textwidth}
\centering
\begin{tikzpicture}
\begin{axis}[
    width=\linewidth, height=5.5cm,
    xlabel={\texttt{nBits}}, ylabel={$R^2$ (split 0)},
    xmode=log, log basis x=2,
    xtick={1024,2048,4096,8192}, xticklabels={1024,2048,4096,8192},
    ymin=0.925, ymax=0.952, grid=both,
    legend style={at={(0.02,0.02)},anchor=south west,font=\scriptsize},
    tick label style={font=\small}, label style={font=\small},
]
\addplot+[mark=*, thick, color=violet!70!black] coordinates {(1024,0.948479) (2048,0.946299) (4096,0.946379)};
\addlegendentry{Radius 2}
\addplot+[mark=square*, thick, color=purple!70!black] coordinates {(1024,0.941717) (2048,0.948305) (4096,0.938020)};
\addlegendentry{Radius 3}
\addplot+[mark=triangle*, thick, color=magenta!55!black] coordinates {(4096,0.933311) (8192,0.930429)};
\addlegendentry{Radii 2+3}
\end{axis}
\end{tikzpicture}
\captionof{figure}{Buchwald-Hartwig split~0: $R^2$ vs.\ \texttt{nBits}.}
\label{fig:bh_r2_vs_nbits_concat}
\end{minipage}
\end{figure*}

\subsubsection{Cross-Split Stability (All Ten Splits)}
Based on the split-0 screening, we select radius~2 and evaluate the three \texttt{nBits} values ($1024$, $2048$, $4096$) across all ten splits. Figures~\ref{fig:sm_all_splits_r2_points} and~\ref{fig:bh_r2_points_nbits} show per-split $R^2$.

On Suzuki-Miyaura, all three \texttt{nBits} settings produce similar mean $R^2$ ($\approx 0.83$-$0.85$), but their variance differs. \texttt{nBits}$\,=2048$ hits a sweet spot: it matches or approaches \texttt{nBits}$\,=4096$ on most splits while training roughly $2.5\times$ faster. \texttt{nBits}$\,=1024$ is fastest but shows slightly more variance, suggesting that hash collisions at lower dimensionality occasionally hurt on certain partitions.

On Buchwald-Hartwig, the three settings are tightly clustered ($R^2 \approx 0.943$-$0.958$), with \texttt{nBits}$\,=4096$ showing marginally lower variance. The weak sensitivity to \texttt{nBits} here reflects the narrower chemical diversity in this combinatorial dataset.

\begin{figure*}[!t]
\centering
\begin{minipage}[t]{0.48\textwidth}
\centering
\begin{tikzpicture}
\begin{axis}[
    width=\linewidth, height=5.5cm,
    xlabel={Split index}, ylabel={$R^2$},
    xmin=-0.2, xmax=9.2, xtick={0,...,9},
    ymin=0.80, ymax=0.89, grid=both,
    legend style={at={(0.02,0.02)},anchor=south west,font=\scriptsize},
    tick label style={font=\small}, label style={font=\small},
]
\addplot+[mark=*, thick, color=violet!75!black] coordinates {
(0,0.829923)(1,0.849640)(2,0.854700)(3,0.836211)(4,0.865449)
(5,0.848952)(6,0.829672)(7,0.849934)(8,0.816335)(9,0.835385)};
\addlegendentry{r=2, 1024}
\addplot+[mark=square*, thick, color=purple!75!black] coordinates {
(0,0.836096)(1,0.861001)(2,0.872920)(3,0.839494)(4,0.862135)
(5,0.843778)(6,0.819766)(7,0.866798)(8,0.845090)(9,0.842700)};
\addlegendentry{r=2, 2048}
\addplot+[mark=triangle*, thick, color=magenta!65!black] coordinates {
(0,0.845056)(1,0.863178)(2,0.866648)(3,0.836394)(4,0.863038)
(5,0.851354)(6,0.846053)(7,0.852099)(8,0.845981)(9,0.840943)};
\addlegendentry{r=2, 4096}
\end{axis}
\end{tikzpicture}
\captionof{figure}{Suzuki-Miyaura: per-split $R^2$ for radius-2 \texttt{nBits} grid.}
\label{fig:sm_all_splits_r2_points}
\end{minipage}\hfill
\begin{minipage}[t]{0.48\textwidth}
\centering
\begin{tikzpicture}
\begin{axis}[
    width=\linewidth, height=5.5cm,
    xlabel={Split index}, ylabel={$R^2$},
    xmin=-0.2, xmax=9.2, xtick={0,...,9},
    ymin=0.940, ymax=0.960, grid=both,
    legend style={at={(0.02,0.02)},anchor=south west,font=\scriptsize},
    tick label style={font=\small}, label style={font=\small},
]
\addplot+[mark=*, thick, color=violet!75!black] coordinates {
(0,0.948479)(1,0.943204)(2,0.946997)(3,0.948450)(4,0.953370)
(5,0.946593)(6,0.945595)(7,0.949148)(8,0.946433)(9,0.944731)};
\addlegendentry{r=2, 1024}
\addplot+[mark=square*, thick, color=purple!75!black] coordinates {
(0,0.946299)(1,0.943916)(2,0.951193)(3,0.949419)(4,0.954338)
(5,0.947515)(6,0.947076)(7,0.951366)(8,0.949630)(9,0.945660)};
\addlegendentry{r=2, 2048}
\addplot+[mark=triangle*, thick, color=magenta!65!black] coordinates {
(0,0.946379)(1,0.946276)(2,0.950591)(3,0.949686)(4,0.950430)
(5,0.948600)(6,0.946296)(7,0.951918)(8,0.957680)(9,0.948497)};
\addlegendentry{r=2, 4096}
\end{axis}
\end{tikzpicture}
\captionof{figure}{Buchwald-Hartwig: per-split $R^2$ for radius-2 \texttt{nBits} grid.}
\label{fig:bh_r2_points_nbits}
\end{minipage}
\end{figure*}

\subsubsection{Radius Ablation at Fixed \texttt{nBits}}
On both datasets, radius~2 consistently achieves the highest or near-highest $R^2$ per split. Radius~3 performs comparably on some splits but is more variable, suggesting that larger neighborhoods pick up less transferable substructural patterns. The concatenated setting underperforms both single-radius alternatives on nearly every split, confirming that the extra dimensionality adds redundancy rather than useful information.

A chemical interpretation of the radius-2 advantage is as follows. For an aryl halide coupling partner, a radius-2 neighborhood centered on the ipso carbon captures the two ortho substituents and their immediate neighbors, encoding the steric and electronic environment most relevant to oxidative addition barriers and Pd coordination geometry. Similarly, for phosphine ligands, radius-2 encompasses the phosphorus center and its three directly bonded substituent groups, which determine the cone angle and donor strength that govern catalytic activity. Extending to radius-3 pulls in para substituents, distal ring atoms, and ligand backbone fragments that vary little across entries in these combinatorial libraries. When folded into a fixed-length vector, these additional substructures increase hash collision frequency without contributing discriminative signal near the reactive center. This is in line with the known sensitivity of cross-coupling yields to ortho substitution patterns and ligand sterics \cite{rogers2010ecfp, yang2019chemprop}.

\begin{figure*}[!t]
\centering
\begin{minipage}[t]{0.48\textwidth}
\centering
\begin{tikzpicture}
\begin{axis}[
    width=\linewidth, height=5.5cm,
    xlabel={Split index}, ylabel={$R^2$},
    xmin=-0.2, xmax=9.2, xtick={0,...,9},
    ymin=0.80, ymax=0.89, grid=both,
    legend style={at={(0.02,0.02)},anchor=south west,font=\scriptsize},
    tick label style={font=\small}, label style={font=\small},
]
\addplot+[mark=*, thick, color=violet!75!black] coordinates {
(0,0.836096)(1,0.861001)(2,0.872920)(3,0.839494)(4,0.862135)
(5,0.843778)(6,0.819766)(7,0.866798)(8,0.845090)(9,0.842700)};
\addlegendentry{r=2}
\addplot+[mark=square*, thick, color=purple!75!black] coordinates {
(0,0.828581)(1,0.846265)(2,0.867324)(3,0.826295)(4,0.850213)
(5,0.839811)(6,0.823697)(7,0.844906)(8,0.840697)(9,0.836338)};
\addlegendentry{r=3}
\addplot+[mark=triangle*, thick, color=magenta!65!black] coordinates {
(0,0.821443)(1,0.825610)(2,0.847035)(3,0.813055)(4,0.854564)
(5,0.824297)(6,0.827167)(7,0.840796)(8,0.837373)(9,0.824244)};
\addlegendentry{r=2+3 concat}
\end{axis}
\end{tikzpicture}
\captionof{figure}{Suzuki-Miyaura: radius ablation at \texttt{nBits}$\,=2048$.}
\label{fig:sm_radius_ablation_points}
\end{minipage}\hfill
\begin{minipage}[t]{0.48\textwidth}
\centering
\begin{tikzpicture}
\begin{axis}[
    width=\linewidth, height=5.5cm,
    xlabel={Split index}, ylabel={$R^2$},
    xmin=-0.2, xmax=9.2, xtick={0,...,9},
    ymin=0.920, ymax=0.956, grid=both,
    legend style={at={(0.02,0.02)},anchor=south west,font=\scriptsize},
    tick label style={font=\small}, label style={font=\small},
]
\addplot+[mark=*, thick, color=violet!75!black] coordinates {
(0,0.946299)(1,0.943916)(2,0.951193)(3,0.949419)(4,0.954338)
(5,0.947515)(6,0.947076)(7,0.951366)(8,0.949630)(9,0.945660)};
\addlegendentry{r=2}
\addplot+[mark=square*, thick, color=purple!75!black] coordinates {
(0,0.948305)(1,0.939307)(2,0.941685)(3,0.940988)(4,0.943641)
(5,0.939082)(6,0.942985)(7,0.943137)(8,0.943359)(9,0.941364)};
\addlegendentry{r=3}
\addplot+[mark=triangle*, thick, color=magenta!65!black] coordinates {
(0,0.933311)(1,0.939649)(2,0.939997)(3,0.926309)(4,0.928619)
(5,0.932226)(6,0.932244)(7,0.938271)(8,0.931370)(9,0.934462)};
\addlegendentry{r=2+3 concat}
\end{axis}
\end{tikzpicture}
\captionof{figure}{Buchwald-Hartwig: radius ablation at \texttt{nBits}$\,=2048$.}
\label{fig:bh_radius_all_splits}
\end{minipage}
\end{figure*}

\subsubsection{Ablation Summary}
Three findings hold across both datasets. First, radius~2 Morgan fingerprints give the strongest and most stable performance; going to radius~3 or concatenating radii does not help and often hurts. Second, within the radius-2 family, \texttt{nBits}$\,=2048$ offers the best balance between accuracy and training time: it approaches the accuracy of larger vectors while training roughly $2.5\times$ faster. Third, Buchwald-Hartwig is less sensitive to fingerprint hyperparameters than Suzuki-Miyaura, reflecting its narrower chemical diversity. For practitioners, a radius-2, \texttt{nBits}$\,=2048$ configuration is a solid default that works well across datasets and splits.

\subsection{Diagnostics and Qualitative Behavior}
\begin{figure*}
   \centering
    \includegraphics[width=1\linewidth]{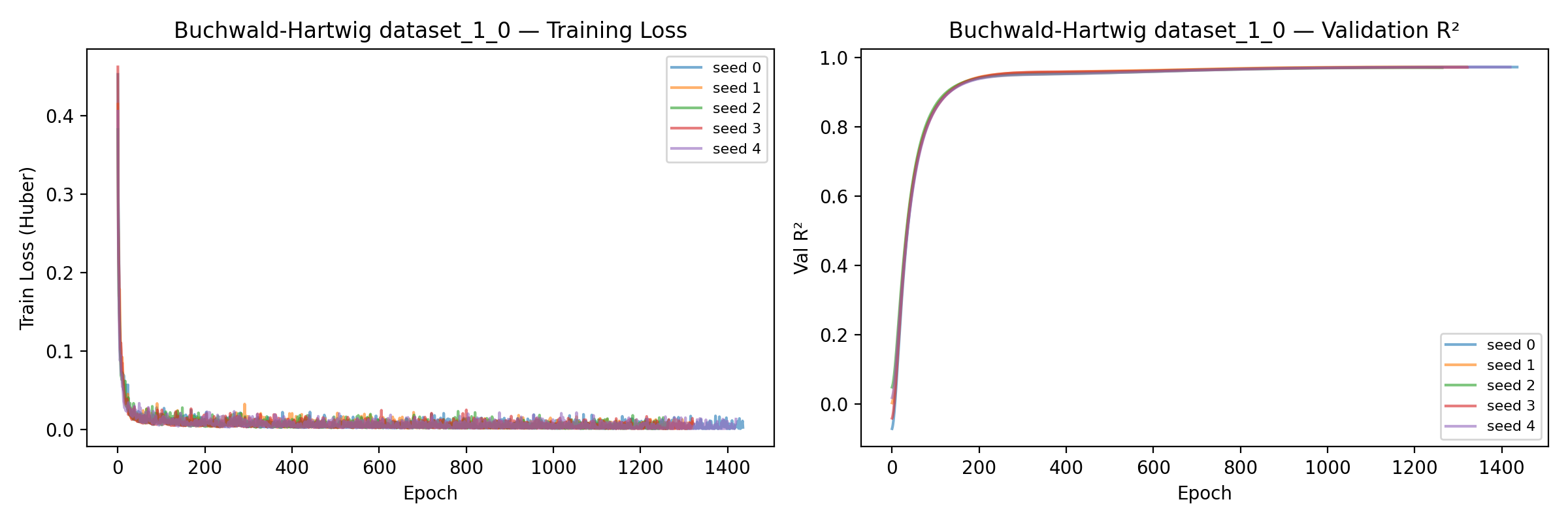}
    \caption{Learning and Validation Curves for Buchwald-Hartwig Dataset}
    \label{fig:ann_baseline_loss}
\end{figure*}

\begin{figure*}
   \centering
    \includegraphics[width=1\linewidth]{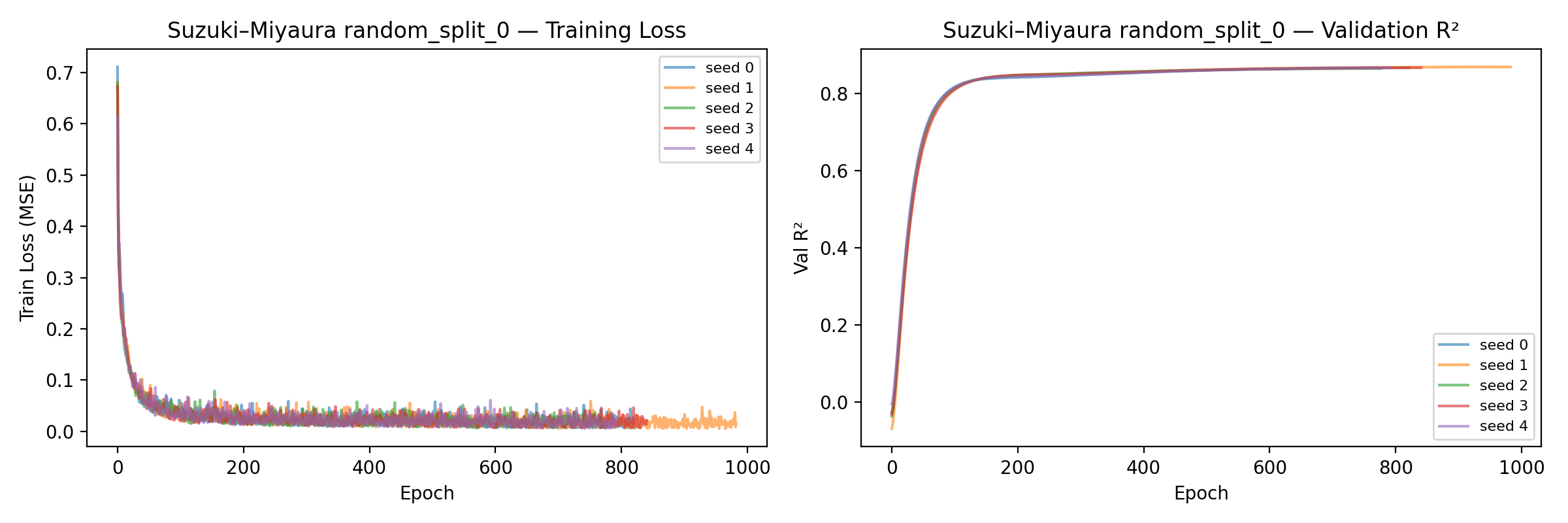}
    \caption{Learning and Validation Curves for Suzuki-Miyaura Dataset}
    \label{fig:morgan_loss}
\end{figure*}

\begin{figure*}[t]
    \centering
    \begin{minipage}[b]{0.5\linewidth}
        \centering
        \includegraphics[width=\linewidth]{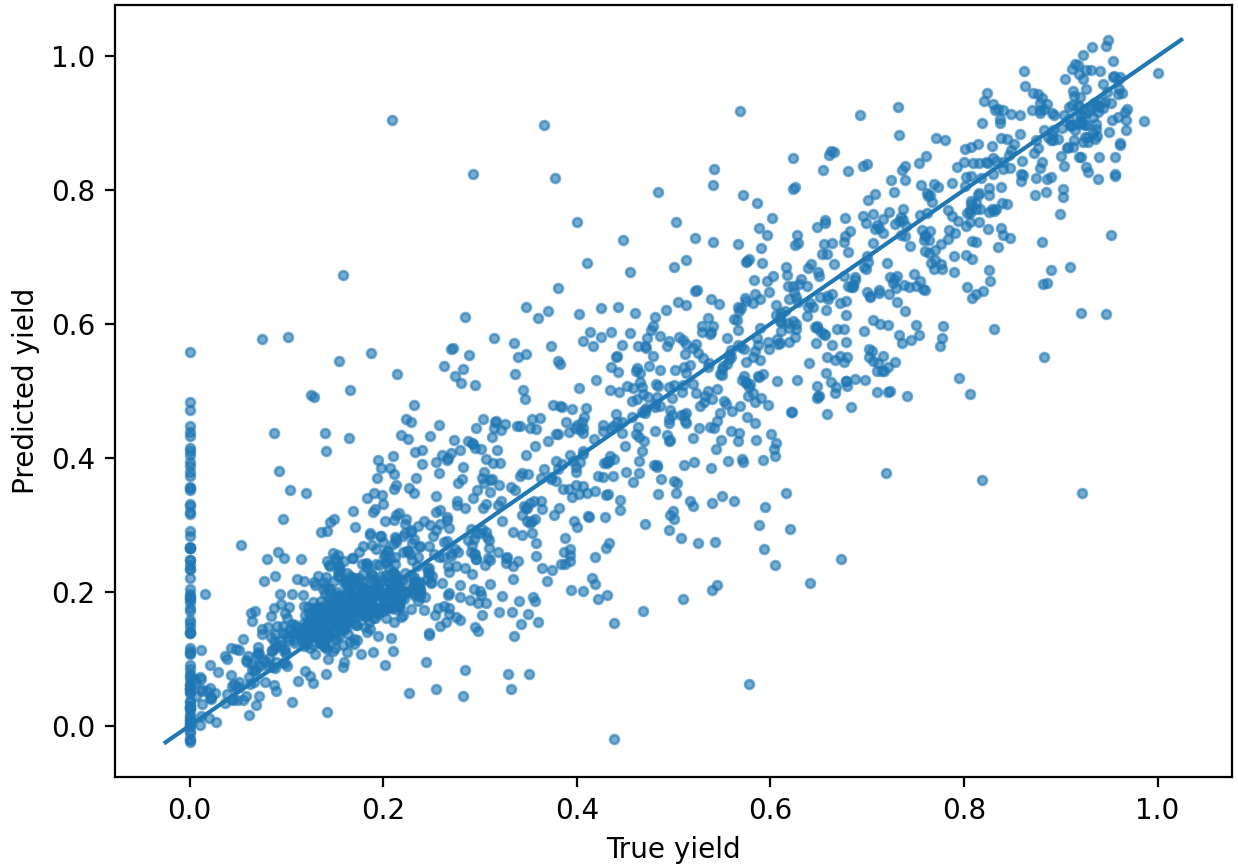}
        \captionof{figure}{True vs Predicted using GNAN}
        \label{fig:gnan_split9}
    \end{minipage}\hfill
    \begin{minipage}[b]{0.5\linewidth}
        \centering
        \includegraphics[width=\linewidth]{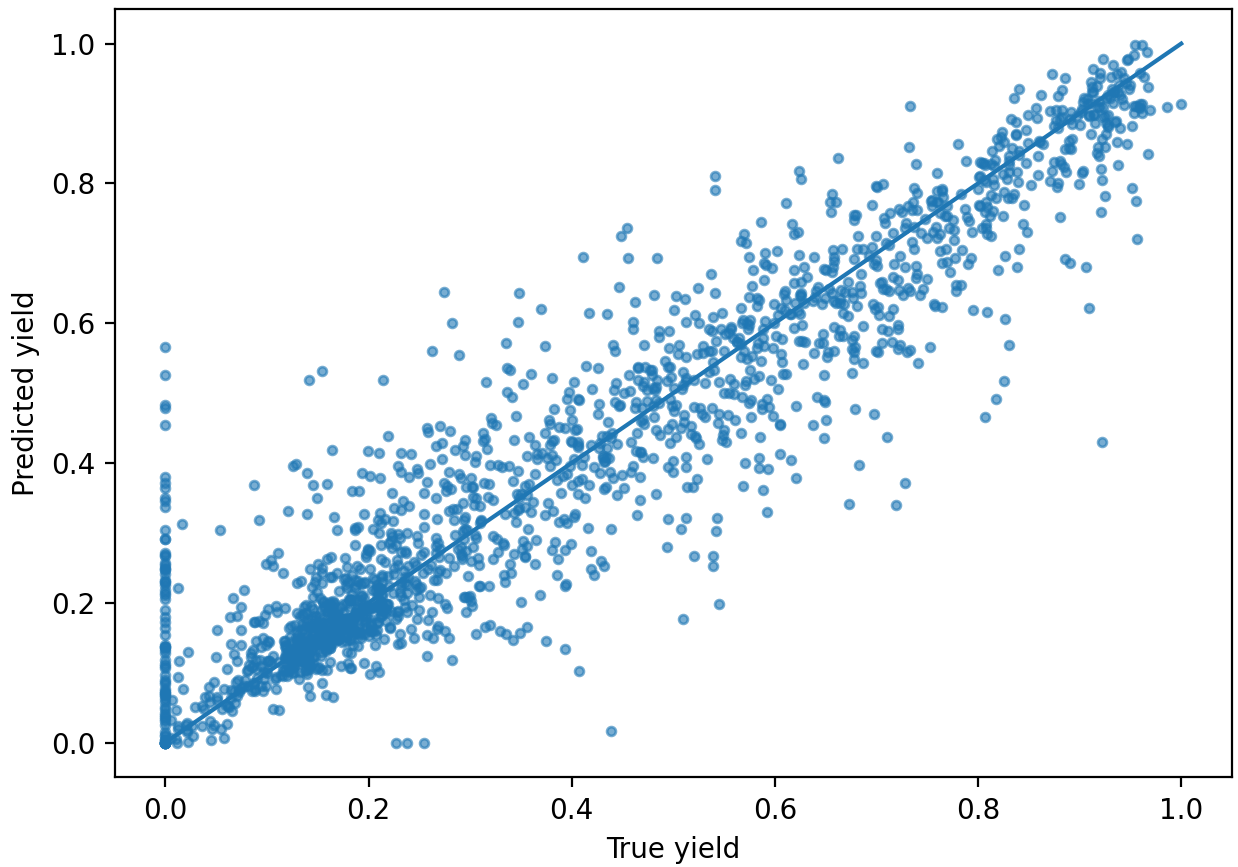}
        \captionof{figure}{True vs Predicted using Morgan FP with ANN}
        \label{fig:ann_baseline_scatter}
    \end{minipage}
\end{figure*}

Learning curves (Figures~\ref{fig:ann_baseline_loss} and \ref{fig:morgan_loss}) confirm that models converge without divergence, indicating stable optimization. Scatter plots (Figures~\ref{fig:ann_baseline_scatter} and \ref{fig:gnan_split9}) show the expected regression pattern: predictions are most accurate in the mid-yield range and regress toward the mean at the extremes. This is expected given label noise and limited coverage of rare component combinations, and matches the challenges documented in analyses of real-world yield data \cite{saebi2023realworld, voinarovska2024challenges}. As Chen \textit{et al.} \cite{ma2024we} pointed out, this pattern partly reflects the yield distribution imbalance: models fit well in the data-rich low-yield region while struggling on sparser high-yield reactions. Addressing this through cost-sensitive reweighting or stratified evaluation is a promising direction for future work.


\section{Limitations and Future Work}
This study reports performance only under random split evaluation. Harder protocols like scaffold splits, component-family holdouts, or temporal partitions would test generalization under distribution shift more rigorously, a concern raised in recent critiques of yield prediction benchmarks \cite{voinarovska2024challenges}. Yield labels are also inherently noisy across sources and conditions, which puts a ceiling on what any method can achieve \cite{saebi2023realworld}. Our ablation uses a single architecture (MLP) and a fixed set of training hyperparameters; the optimal fingerprint configuration might differ with a different regressor or regularization scheme. Future work will (i) add harder split protocols to test out-of-distribution robustness, (ii) explore hybrid architectures combining fingerprint efficiency with the representational power of learned graph or sequence embeddings \cite{han2024pretrained_gnn, shi2024reamvp}, and (iii) investigate cost-sensitive training to improve predictions on underrepresented high-yield reactions \cite{ma2024we}. Combining uncertainty-aware methods like deep kernel learning \cite{singh2024deep} or GNAN's heteroscedastic regression \cite{kwon2022gnan} with lightweight fingerprint representations is another avenue worth exploring.

\section{Conclusion}
We introduced MFP, a reaction yield prediction method that
encodes each reaction as a fixed-length descriptor from
role-aware Morgan fingerprints. Six fingerprint blocks capture
component identity, structural transformation, and
transformation magnitude, giving a lightweight MLP enough
signal to approach GNAN accuracy ($R^2 = 0.969$ on
Buchwald-Hartwig, $R^2 = 0.878$ on Suzuki-Miyaura) while
training an order of magnitude faster. A formal complexity
analysis confirmed that this speed advantage is structural:
MFP pays the representation cost once, while graph methods
repeat it at every training step. A paired Wilcoxon test
showed the gap to GNAN is statistically significant
($p \approx 0.02$), placing MFP between GNAN and YieldBERT.
Feature block ablation revealed high inter-block redundancy:
no single block is indispensable, and the model degrades
gracefully when any block is removed.

Ablation over radius and folded vector length identified
radius-2 at \texttt{nBits}$\,=2048$ as the strongest default.
Radius-2 neighborhoods capture the ortho substitution and
ligand steric environments that drive cross-coupling yield,
while radius-3 adds conserved scaffold fragments that increase
hash collisions without improving discrimination. We release
all code and trained models at
\url{https://github.com/chinmaymirji/morgan-fp-yield-prediction.git}
and recommend this configuration as a practical baseline
against which more complex architectures can be measured.

\section*{Acknowledgment}
The authors thank the open-source contributors and dataset authors whose resources enabled this study, including the YieldBERT resources \cite{schwaller2021yieldbert} and the Buchwald-Hartwig dataset \cite{ahneman2018buchwald_ml}.

\bibliographystyle{IEEEtran}
\bibliography{cdse}

\end{document}